\documentclass[conference]{IEEEtran}
\usepackage[utf8]{inputenc}
\IEEEoverridecommandlockouts
\usepackage{cite}
\usepackage{amsmath,amssymb,amsfonts}
\usepackage{algorithmic}
\usepackage{graphicx}
\usepackage{float}
\usepackage{textcomp}
\usepackage{hyperref }
\usepackage{epstopdf}
\def\BibTeX{{\rm B\kern-.05em{\sc i\kern-.025em b}\kern-.08em
    T\kern-.1667em\lower.7ex\hbox{E}\kern-.125emX}}
    
\begin{document}

\title{Intracranial Error Detection via Deep Learning\\
\thanks{This work was supported by DFG grant EXC1086 BrainLinks-BrainTools, Baden-Württemberg Stiftung grant BMI-Bot, Graduate School of Robotics in Freiburg, Germany and the State Graduate Funding Program of Baden-Württemberg, Germany.}
}

\author{

    \IEEEauthorblockN{Martin Völker}
    \IEEEauthorblockA{\textit{Graduate School of Robotics}\\
    \textit{Albert-Ludwigs-University}\\
    Freiburg, Germany\\
    martin.voelker@uniklinik-freiburg.de}
    
    \and
    
    \IEEEauthorblockN{Jiří Hammer}
    \IEEEauthorblockA{\textit{Department of Neurology}\\
    \textit{Motol University Hospital, Charles University}\\
    Prague, Czech Republic\\
    jiri.hammer@lfmotol.cuni.cz}

    \and
    
    \IEEEauthorblockN{Robin T. Schirrmeister}
    \IEEEauthorblockA{\textit{Translational Neurotechnology Lab}\\
    \textit{University Medical Center Freiburg}\\
    Freiburg, Germany\\
    robin.schirrmeister@uniklinik-freiburg.de}
    
    \and
    
    \IEEEauthorblockN{Joos Behncke}
    \IEEEauthorblockA{\textit{Department of Computer Science}\\
    \textit{Albert-Ludwigs-University}\\
    Freiburg, Germany\\
    joos.behncke@uniklinik-freiburg.de}
    
    \and
    
    \IEEEauthorblockN{Lukas D.J. Fiederer}
    \IEEEauthorblockA{\textit{Faculty of Biology}\\
    \textit{Albert-Ludwigs-University}\\
    Freiburg, Germany\\
    lukas.fiederer@uniklinik-freiburg.de}
    
    \and
    
    \IEEEauthorblockN{Andreas Schulze-Bonhage}
    \IEEEauthorblockA{\textit{Epilepsy Center}\\
    \textit{University Medical Center Freiburg}\\
    Freiburg, Germany\\
    andreas.schulze-bonhage@uniklinik-freiburg.de}
    
    \and
    
    \IEEEauthorblockN{Petr Marusič}
    \IEEEauthorblockA{\textit{Department of Neurology}\\
    \textit{Motol University Hospital, Charles University}\\
    Prague, Czech Republic\\
    petr.marusic@lfmotol.cuni.cz}
    
    \and

    \IEEEauthorblockN{Wolfram Burgard}
    \IEEEauthorblockA{\textit{Department of Computer Science}\\
    \textit{Albert-Ludwigs-University}\\
    Freiburg, Germany\\
    burgard@informatik.uni-freiburg.de}
    
    \and
    
    \IEEEauthorblockN{Tonio Ball}
    \IEEEauthorblockA{\textit{Translational Neurotechnology Lab}\\
    \textit{University Medical Center Freiburg}\\
    Freiburg, Germany\\
    tonio.ball@uniklinik-freiburg.de}
}

\maketitle

\begin{abstract}
Deep learning techniques have revolutionized the field of machine learning and were recently successfully applied to various classification problems in noninvasive electroencephalography (EEG). However, these methods were so far only rarely evaluated for use in intracranial EEG. We employed convolutional neural networks (CNNs) to classify and characterize the error-related brain response as measured in 24 intracranial EEG recordings. Decoding accuracies of CNNs were significantly higher than those of a regularized linear discriminant analysis. Using time-resolved deep decoding, it was possible to classify errors in various regions in the human brain, and further to decode errors over 200\,ms before the actual erroneous button press, e.g., in the precentral gyrus. Moreover, deeper networks performed better than shallower networks in distinguishing correct from error trials in all-channel decoding. In single recordings, up to 100\,\% decoding accuracy was achieved. Visualization of the networks' learned features indicated that multivariate decoding on an ensemble of channels yields related, albeit non-redundant information compared to single-channel decoding. In summary, here we show the usefulness of deep learning for both intracranial error decoding and mapping of the spatio-temporal structure of the human error processing network.

\end{abstract}


\section{Introduction}
Neurotechnological applications such as brain-computer interfaces (BCIs) can be improved by error decoding in electroencephalography (EEG)\cite{spuler2012online,iturrate2013shared,chavarriaga2015decoding,kreilinger2012error, dias2018masked}. In addition to noninvasive recording techniques, intracranial EEG was also shown to be useable for error decoding \cite{milekovic2013detection,wander2013cortically,even2017augmenting}.

However, when it comes to real-life applications, a high accuracy is decisive for the usefulness of such error decoding. In recent years, deep learning has driven the state-of-the-art of decoding accuracies in various fields of research \cite{lecun2015deep, goodfellow2016deep}, especially in computer vision \cite{krizhevsky2012imagenet} and speech recognition \cite{amodei2016deep} or generation \cite{van2016wavenet}.
Newly, deep learning techniques have also been successfully applied to an increasing number of decoding problems in EEG \cite{Schirrmeister:2017bv,Behncke:2017ug,schirrmeister2017pathology, putten2018predicting,sors2018convolutional} and further utilized for extraction and visualization of learned features \cite{hartmann2018hierarchical}. We previously reported that convolutional neural networks (CNNs) performed better than regularized linear discriminant analysis (rLDA) and filter bank common spatial patterns (FBCSP) algorithms in error decoding from noninvasive EEG \cite{volker2017between}, and can be used to reliably classify errors in inter-subject decoding \cite{volker2018deep}.
From different CNN architectures, residual neural networks (ResNets) are a particularly promising architecture for challenging classification problems. ResNets employ a specialized CNN architecture with a typically very large number of convolutional layers \cite{he2016deep}, and have only recently been applied for classification in noninvasive EEG \cite{Schirrmeister:2017bv, wang2018ResNet}.
Deep neural networks have also been successfully applied in the classification or prediction of epileptic signals \cite{krug2008cnn, antoniades2016deep, ahmedt2018deep, antoniades2017detection, hosseini2017optimized} and movements \cite{xie2018decoding, wang2017ajile}. Other than that, deep learning techniques for BCIs based on intracranial recordings are so far mostly unexplored.
Here we show for the first time the usefulness of CNNs and ResNets for both intracranial error decoding and as a tool for precise spatio-temporal brain mapping.

\section{Experiment}
In Freiburg, Germany, and in Prague, Czech Republic, 23 patients who were implanted with intracranial electrodes due to pharmacoresistant epilepsy participated in a flanker task experiment as described in \cite{volker2018dynamics}. One patient participated in two recording sessions on different days with a different subset of the intracranial electrodes; thus, 24 recording sessions were available overall. All patients gave their written informed consent before participating in the study. The study was approved by the local Ethics Committees. Trials were epoched on the onset of the correct or erroneous movement with the left or right index finger, as measured by analog joystick buttons. On average, patients executed 212 ± 12 (mean ± SEM) correct and 51 ± 5 error trials and had an error rate of 19.34 ± 1.79\,\%.

\section{Preprocessing, Decoding \& Statistics}
Locations of the stereotactic depth electrodes were identified with the help of post-implantation MRI or CT as in \cite{pistohl2012decoding} and transformed into the MNI coordinate space \cite{evans1992anatomical}. Each electrode was then assigned to a specific brain region by calculating cytoarchitectonic probabilistic maps in the SPM anatomy toolbox \cite{eickhoff2005new}. Intracranial EEG data were re-referenced bipolarly between the respective neighbors to be specific for local effects and reduce external noise contamination, and resampled to 250 Hz. Other than that, the EEG data were only minimally pre-processed, as described in \cite{volker2018deep}, to operate under application-oriented conditions. We used open-source python implementations for both rLDA \cite{ledoit2004well} and CNN classifiers. Deep4Net and ShallowNet architectures were employed as described in \cite{Schirrmeister:2017bv} and available in the Braindecode Toolbox\footnote{\url{https://robintibor.github.io/braindecode/source/braindecode.models.html}}. Additionally, we used a 34-layered ResNet architecture\footnote{\url{https://github.com/robintibor/adamw-eeg-eval/blob/445cc5d471d8eea3814ffa39621974dda7c471a6/adamweegeval/resnet.py}} and the compact EEGNet architecture was reimplemented as described in \cite{lawhern2016eegnet}, as the EEGNet code from the original publication was not available. As optimizer, we used AdamW \cite{loshchilov2017fixing} with cosine annealing \cite{loshchilov2016sgdr}, a weight decay of 0.002 and an initial learning rate of $\frac{0.01}{32}$. For each recording, the first 60\,\% of the data was used for training, and the last 40\,\% were reserved as final evaluation set, which was only used to test the final accuracies.

Statistical significance of the single-channel classifications was evaluated by randomly permuting the true labels of the test set $10^6$ times to generate a null distribution. For significance of brain region accuracy averages and classifier comparisons, a Wilcoxon signed rank test was employed \cite{wilcoxon1945individual}.

The classification of errors is typically a problem with a strong trial imbalance, as correct trials occur far more often in realistic applications. 
Error decoding studies also usually report a higher classification accuracy for the correct class \cite{chavarriaga2014errare}. In a decoding problem with such a strongly imbalanced number of trials per class, it is thus advisable to define a normalized accuracy acc\textsubscript{norm} as:
\begin{equation}
acc\textsubscript{norm} = \frac{1}{nClasses} *  \sum\limits_{i=1}^{nClasses}TPR(class\textsubscript{i})
\end{equation}

with the True Positive Rate (TPR or sensitivity\cite{altman1994diagnostic}) as:

\begin{equation}
TPR(class\textsubscript{i}) = \frac{TP(class\textsubscript{i}) }{TP(class\textsubscript{i})  + FN(class\textsubscript{i}) }
\end{equation}

TP is the number of true positives, and FN the number of false negative decoding results per class. Thus, when speaking of acc\textsubscript{norm} in the context of our binary error decoding problem, we define it as arithmetic mean of TPR(correct) and TPR(error); this method is also known as macro-averaging \cite{lewis1992evaluation}. Macro-averaging prohibits classifiers from achieving seemingly high accuracies by exploiting the trial imbalance, e.g., by only predicting the most abundant class for all trials, as the chance level of acc\textsubscript{norm} always stays at \textbf{$\frac{1}{nClasses}$}. To get an idea of the number of true negatives (TN) and false positives (FP), we also included the specificity (or true negative rate),

\begin{equation}
specificity(class\textsubscript{i}) = \frac{TN(class\textsubscript{i}) }{TN(class\textsubscript{i})  + FP(class\textsubscript{i}) }
\end{equation}

as well as the F1 score, i.e., the harmonic mean of precision and sensitivity, as further measures.
\begin{equation}
F1(class\textsubscript{i}) = \frac{2*TP(class\textsubscript{i}) }{2*TP(class\textsubscript{i})  + FP(class\textsubscript{i}) + FN(class\textsubscript{i}) }
\end{equation}

To cope with the class imbalance, we further used Braindecode's \textit{ClassBalancedBatchSizeIterator}, which draws the training samples such that, in expectation, the same number of examples is drawn per class. Other metrics, like training accuracies, were also macro-averaged.

\section{Comparison of Classifiers}

\subsection{Single-Channel Decoding}
Channel-selection in intracranial EEG is not trivial, especially as the electrode locations are unique in every measurement and do not follow a certain spatial pattern as is usually the case in noninvasive EEG. Therefore, we first compared the classifiers methods in single-channel decoding. Table \ref{tabClassifierComparison} gives a comparison of classifier performance in this setting, evaluated with 200 training epochs.

\begin{figure*}
\centerline{\includegraphics[width=1.0\textwidth]{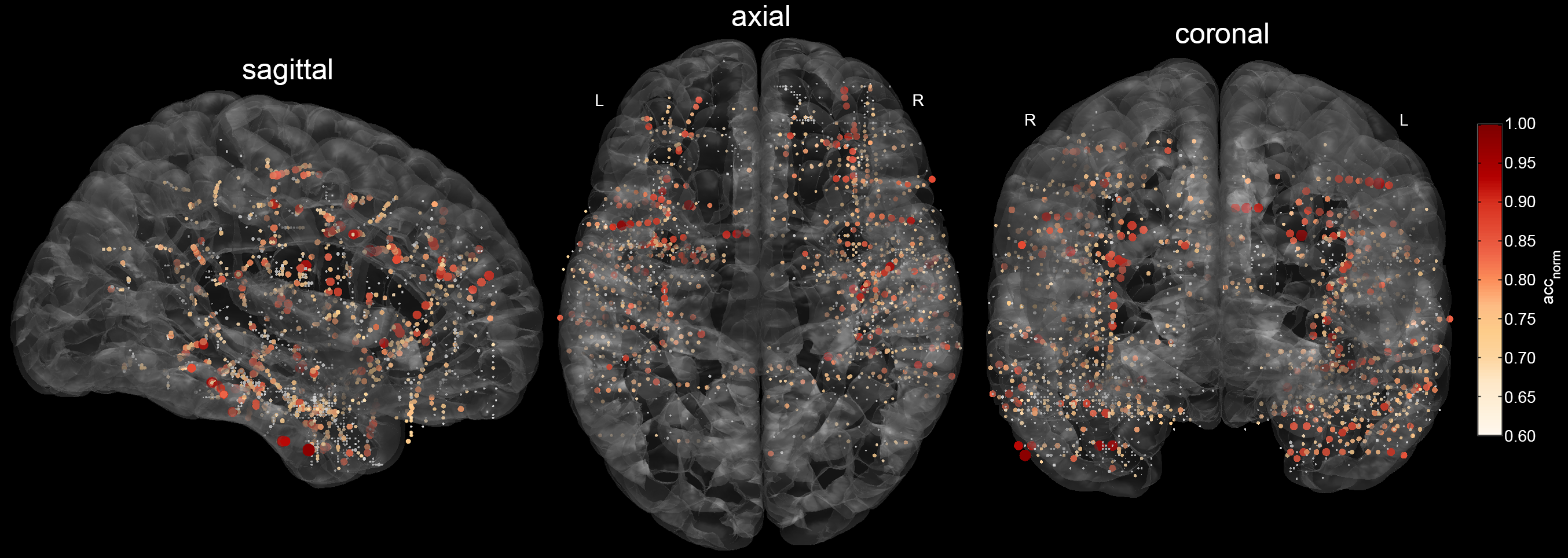}}
\caption{Single-channel decoding using the Deep4Net on a 2s-window (-0.5s to 1.5s in regard to the button press event) of intracranial EEG data. All 2332 channels are marked according their MNI coordinates in an ICBM152 brain template \cite{mazziotta2001probabilistic}; normalized accuracies $>$ 60\,\% are plotted color- and size-coded.}
\label{fig3Dbrain}
\end{figure*}

\begin{figure*}[h]
\centering{\includegraphics[width=0.94\textwidth]{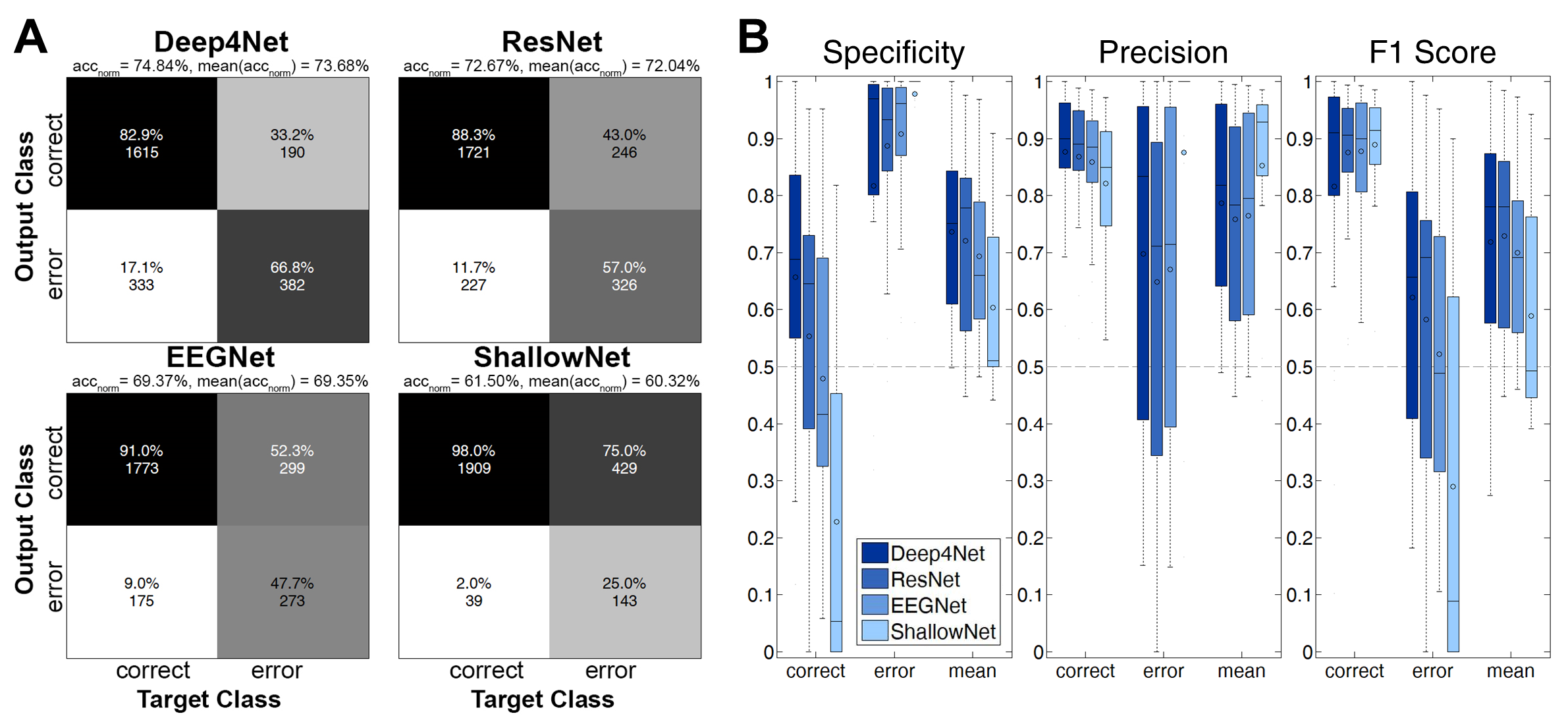}}
\caption{Classifier performance in all-channel decoding. Here, the classifiers were trained on all available channels per patient. A) Confusion matrices of the four models used for decoding. The matrices display the sum of all trials over the 24 recordings. On top of the matrices, the normalized accuracy over all trials, i.e., acc\textsubscript{norm}, and the mean of the single recordings' normalized accuracy, i.e., mean (acc\textsubscript{norm}) is displayed; please note that these two measures differ slightly, as the patients had a varying number of total trials and trials per class. B) Box plots for specificity, precision and F1 score. The box represents the interquartile range (IQR) of the data, the circle within the mean, the horizontal line depicts the median. The lower whiskers include all data points that have the minimal value of \textit{$25^{th}$percentile-1.5*IQR}, the upper whiskers include all points that are maximally \textit{$75^{th}$percentile+1.5*IQR}. }
\label{figEndtoEndComparison}
\end{figure*}

\begin{table}[h]
\caption{Classifier performance in single-channel decoding. For both single-class and normalized accuracy, mean ± sem are listed.}
\begin{center}
\begin{tabular}{|c|c|c|c|}
\hline
\textbf{Classifier}&\textbf{acc\textsubscript{norm} (\%)}&\textbf{acc\textsubscript{correct} (\%)}&\textbf{acc\textsubscript{error} (\%)} \\
\hline
Deep4Net & 59.28 ± 0.50 & 69.37 ± 0.44 & 49.19 ± 0.56 \\
ShallowNet & 58.42 ± 0.32 & 74.83 ± 0.25 & 42.01 ± 0.40 \\ 
EEGNet & 57.73 ± 0.52 & 57.78 ± 0.48 & 57.68 ± 0.56 \\ 
rLDA & 53.76 ± 0.32 & 76.12 ± 0.26 & 31.40 ± 0.38 \\ 
ResNet & 52.45 ± 0.21 & 95.47 ± 0.14 & 09.43 ± 0.28 \\ 
\hline
\end{tabular}
\label{tabClassifierComparison}
\end{center}
\end{table}

In single-channel decoding, the Deep4Net had the best normalized accuracy. This difference was significant with p$<$0.001 (Wilcoxon signed-rank test) in regard to all other methods. For the Deep4Net, we thus visualized the single channel accuracies in Fig. \ref{fig3Dbrain}. The broad distribution of electrodes with high accuracies indicates that errors were not only decodable from limited brain regions; rather, there seems to be an extended error-processing network, and the CNN was able to classify errors in multiple nodes within this network. All in all, electrodes in central and frontal regions had higher accuracies than electrodes in parietal locations.

\begin{figure*}[h]
\centering{\includegraphics[width=0.90\textwidth]{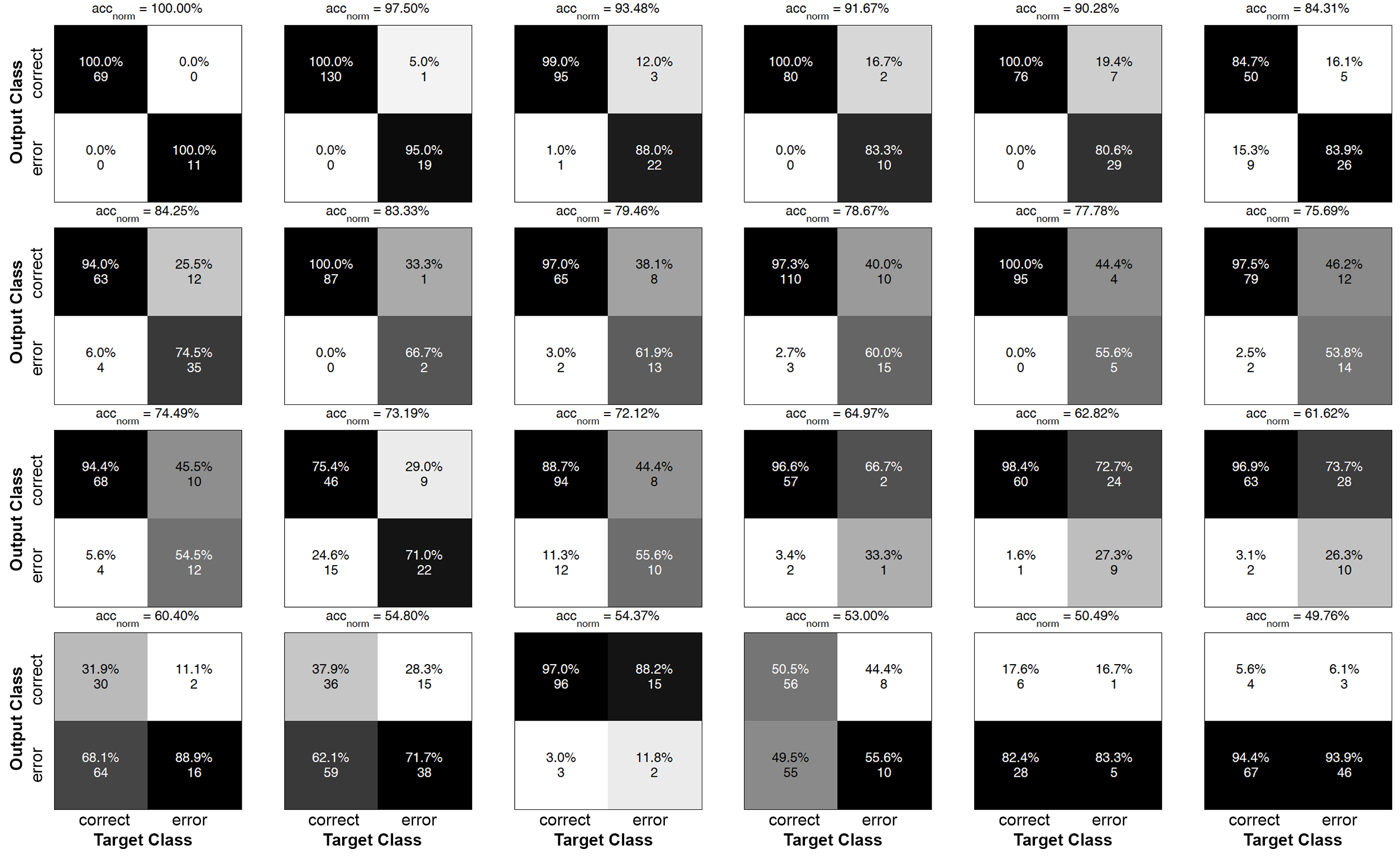}}
\caption{Single-recording confusion matrices of the Deep4 network performing all-channel error decoding. The matrices are sorted in a descending order according to the normalized decoding accuracy. On top of each confusion matrix, the corresponding normalized accuracy is specified.}
\label{figConfusionMatricesDeep4}
\end{figure*} 

\subsection{Decoding Using All Channels}
As individual channels may carry non-redundant information, multivariate decoding from all available channels simultaneously may increase decoding performance. We compared network architectures in this application scenario (Fig. \ref{figEndtoEndComparison} A) using 1000 training epochs.
Here, the Deep4Net performed best with an normalized accuracy of 74.84\,\%, ResNet (72.67\,\%), and EEGNet (69.37\,\%) followed, while the ShallowNet was far behind (61.50\,\%). Measures of specificity, precision, and the F1 score per class (Fig. \ref{figEndtoEndComparison} B), indicate that especially the two smaller networks (i.e., networks with fewer parameters, EEGNet and ShallowNet) tended to overpredict the "correct" class, leading to low sensitivity for the error class. The average F1 score for the ShallowNet was significantly (p$<$0.01, Wilcoxon  signed-rank  test) lower than that for the other 3 architectures.
We plotted the confusion matrices of the single recordings for the Deep4Net in Fig. \ref{figConfusionMatricesDeep4}. While there were recordings in which the CNN performed with 100\,\% or very close to perfect accuracy, there were also some in which the classifier predicted nearly only one class and thus performed with an acc\textsubscript{norm} around 50\,\%.
There could be multiple reasons for that. First, as the location of the electrodes is determined by the epilepsy focus, it could be that those recordings did not include any useful locations for error decoding. Second, it could be that a strong class imbalance led to an overfitting on one class. Third, some recordings had very little training data, and one could assume that the CNN was not able to learn the class representations reliably.
However, using Spearman's rank correlation, the decoding performance did neither significantly correlate with the number of errors (r=0.23, p=0.28), the correct trials in the training data (r=-0.07, p=0.75), nor the total error rate (r=-0.05, p=0.82).

\begin{figure}
\centering{\includegraphics[width=0.45\textwidth]{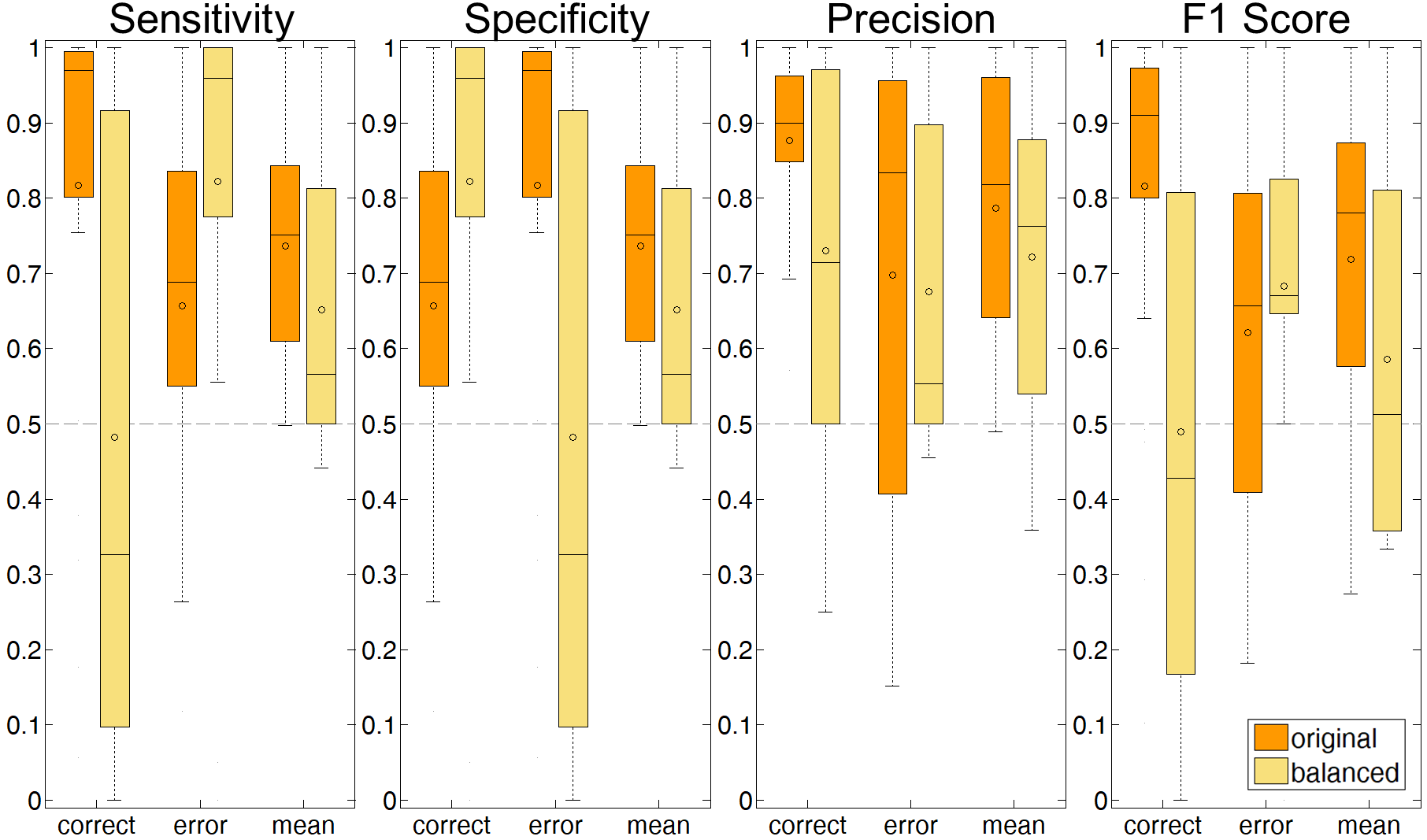}}
\caption{All-channel decoding (Deep4Net) on the original, imbalanced data and a balanced dataset using sub-sampling of the more frequent class.}
\label{figBalanced}
\end{figure}

Generally, it is a known problem in the field of error decoding that classifiers have a higher sensitivity on the correct class. It is believed that the strong trial imbalance could be one reason for this issue \cite{chavarriaga2014errare}.
We used a class-balanced batch size iterator to reduce this problem; but as this method can repeatedly choose samples of the less frequent class, it could theoretically still overfit on this class.
To test for that, we repeated the classification with the Deep4Net, while keeping the number of trials of each class in training and test set equal (Fig. \ref{figBalanced}). For each error trial, the nearest correct trial (randomly before or after) was chosen to keep. In that way, any changes over the time of the experiment could not influence the decoding; otherwise the CNN might mistakenly associate unrelated slow changes in the EEG signal with the class that was more prominent during that time of the recording.

Interestingly, after trial balancing, the sensitivity for the correct class decreased distinctly, while the sensitivity for the error class increased. However, the precision did not improve for the error class, while it decreased for the correct class. The F1 score of the correct class decreased strongly, while the error class saw only little improvement. The average F1 score decreased. All in all, it does not seem to be advisable to balance the number of trials by sub-sampling.

\begin{figure*}[p!]
\centerline{\includegraphics[width=\textwidth]{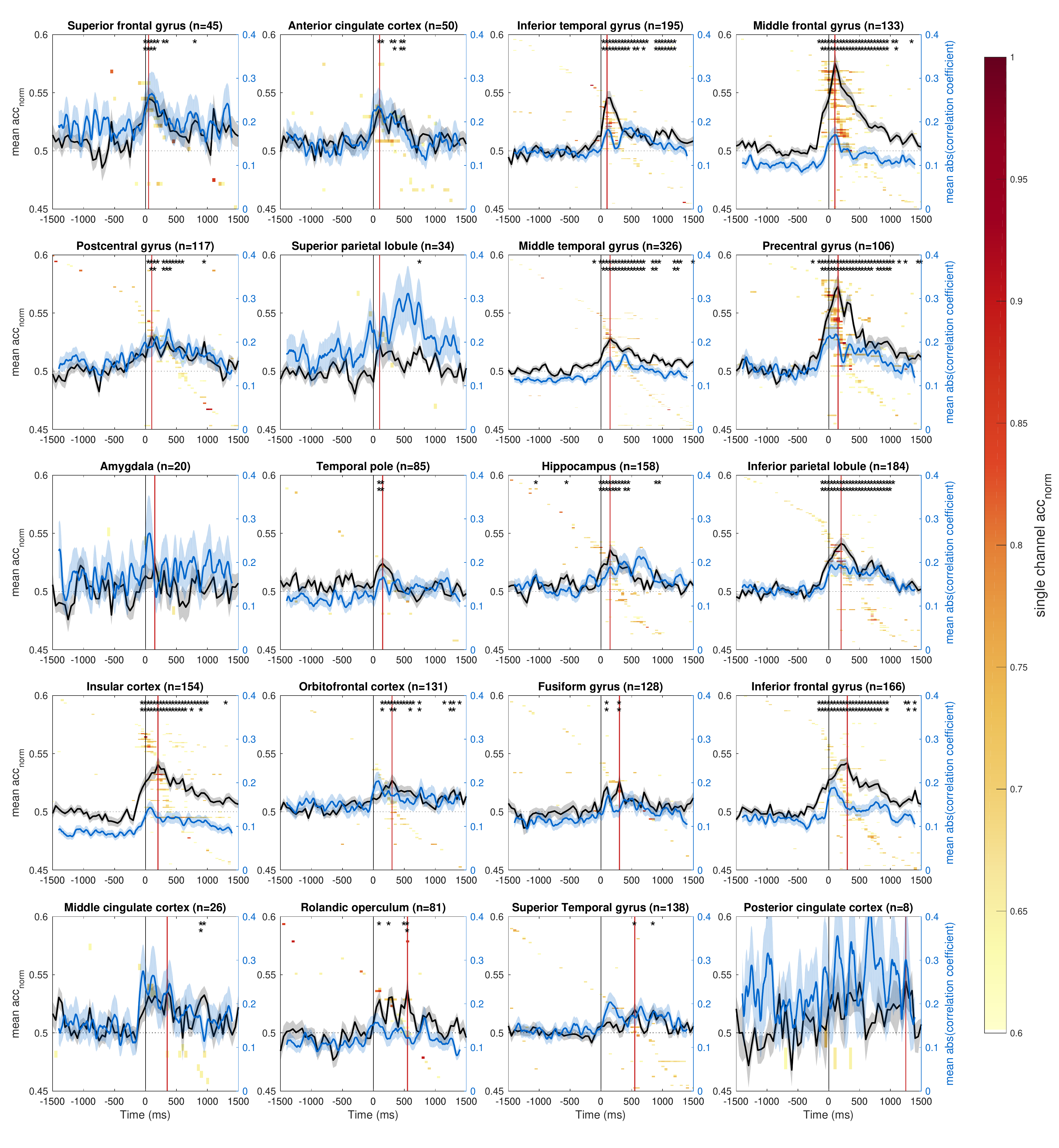}}
\caption{Single-channel decoding using the Deep4Net in a time-resolved manner. Per area, significant (p$<$0.01, permutation test) single-channel accuracies are plotted color-coded in the background. The black curve displays the mean normalized decoding accuracy in the respective area, as calculated within a 200\,ms sliding window (50\,ms step size), and the standard error of the mean (SEM) is plotted semitransparent. Time points with accuracies significantly higher than the the 50\,\% chance level are highlighted by asterisks (* = p$<$0.001, ** = p$<$0.0001, Wilcoxon signed-rank test). The plots were sorted according to the time point of the highest mean accuracy per area (red line). The blue curve represents a visualization of the CNN's learned attributes, as calculated by time-resolved voltage feature input-perturbation network-prediction correlation mapping. Analog to the 200\,ms decoding window, we applied a moving-average filter with a 200\,ms gaussian window on the correlation values.}
\label{figAccsTimeResolved}
\end{figure*}

\begin{figure*}[h]
\centering{\includegraphics[width=\textwidth]{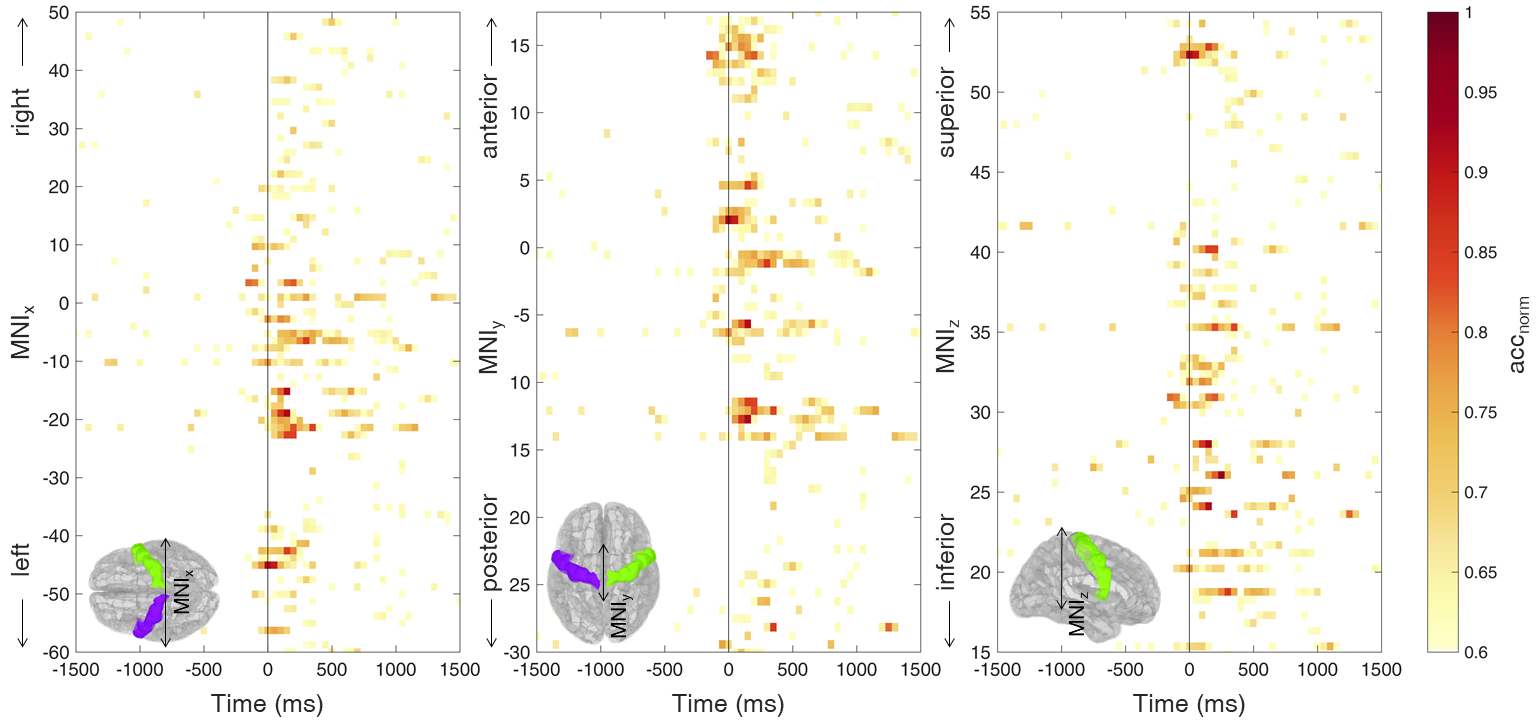}}
\caption{Time-resolved normalized decoding accuracy in the precentral gyrus in relation to the spatial distribution of channels in 3D space. In the bottom left corners, the precentral gyrus was visualized with Brainstorm \cite{tadel2011brainstorm} with the respective orientation (left side purple, right side green).}
\label{figPrecentral}
\end{figure*}

\section{Visualizing the spatio-temporal properties of the error response}

To further dissect the temporal evolution of the error-related response, we used a shifting 200\,ms window on each channel for decoding with the Deep4Net. Due to the short time window, we had to deviate here from the default network parameters and use a stride of 2 samples, as well as reduced filter time lengths of 2 samples.
Moreover, time-resolved voltage feature input-perturbation network-prediction correlations were calculated to visualize the network's learned features (detailed description in \cite{Schirrmeister:2017bv,hartmann2018hierarchical}).

Importantly, the decoding results were obtained via single-channel decoding, while the input-perturbation network-prediction correlations were calculated on a model trained on all channels of the respective patient. To ensure a proper comparison of the two methods, we used the same network parameters as in the shifting-window classification for this analysis. Results from electrodes in each of 20 regions of interest (ROIs) were then pooled and illustrated in Fig. \ref{figAccsTimeResolved}.

In the precentral gyrus, it was possible to significantly (p$<$0.001 in the region mean) decode errors 250\,ms (window center) before the actual button-press event; In the middle and inferior frontal gyrus, errors were significantly decodable 150\,ms before the event. Activity in the hippocampus could even be predictive 1\,s and 0.5\,s before an error.

Generally, the peak of maximal decodability shifted from frontal to parietal and temporal brain regions over the time course of the error response.
The reduced stride and filter time lengths resulted in lower total decoding accuracies; nevertheless, for comparison of the spatio-temporal distribution of the error response, this proved to be a useful method.

Input-perturbation network-prediction correlations (Fig. \ref{figAccsTimeResolved}, blue curves) and time-resolved decoding accuracies shared a similar development over time, although there are some differences, e.g., in the time point of the peak and sharpess of the slope, as seen in the inferior frontal gyrus or in the superior temporal gyrus.

In contrast to the single-channel decoding accuracies, the highest normalized network-prediction correlation values were found at parietal sites, such as the superior parietal lobule. This could indicate that parietal sites rely more on large-scale network-based activity during error processing, which is not easily decodable from single channels.

Similar to the accuracies' development over time, activities at frontal and precentral areas, such as the precentral gyrus and the middle frontal gyrus, were the first to exhibit a sharp increase of the correlation values before the actual error event, indicating that the network learned these to be predictive to the event outcome.

As the precentral gyrus was the most promising area for the detection of errors prior to the erroneous event, we further sorted the single channel accuracies according to their spatial distribution in the MNI space, to examine which exact parts of the precentral gyrus are activated earliest (Fig. \ref{figPrecentral}).

Distinctively, activity in the anterior parts of the precentral gyrus was predictive for errors earlier than in the posterior part. This fits the observation that frontal regions were also activated slightly before the error event. Moreover, central parts of the precentral gyrus were predictive earlier than lateral (left and right) regions. This hints toward a central origin of the error response, which is then distributed to lateral parts of the motor system.

\section{Discussion}

\subsection{Classifier Comparison}
When decoding on all channels, especially deeper networks strongly gained accuracy compared to single-channel decoding. Notably, the ShallowNet performed second-best on single-channel decoding and last in all-channel decoding, while the ResNet, which is the deepest of the tested architectures with 34 convolutional layers, performed worst in single-channel decoding, and second-best in all-channel decoding on all channels. One could thus speculate that deeper networks need a broader spectrum of data to perform reliably, while very shallow networks cannot cope with the amount of noise in all-channel decoding. The 4-layered Deep4Net, however, seems to be a good compromise, as it performed best in both modalities.

\subsection{Error Detection \& Prediction}
We have shown that CNNs are able to detect errors on a single-trial basis in intracranial recordings from precentral and frontal brain regions even before the actual error event, i.e., the false button press, happened. Especially activity in the precentral gyrus was predictive for errors at early time points. Even earlier decodability in the hippocampus could hint toward a role of memory recall for the task performance.

The predictability of errors is an active topic in many areas of neuroimaging; e.g., error-prone patterns were identified in fMRI \cite{eichele2008prediction}, preceding errors up to 30s in the default mode network (DMN), especially in the medial prefrontal cortex (mPFC).
There, preparatory action in a prefrontal–extrastriate network is assumed to happen. Moreover, the contingent negative variation (CNV), an event-related slow wave related to preparatory attention \cite{tecce1972contingent} was shown to be reduced from 100\,ms before an actual error \cite{padilla2006lapses} and is assumed to be partly generated in the pre-supplementary motor area (pre-SMA). It has also been shown that the DMN can be mapped using intracranial measurements in humans, and that precentral and midfrontal regions are an important part of the human DMN \cite{miller2009direct}.
In noninvasive EEG, activity at frontocentral scalp sites proximal to the anterior cingulate cortex (ACC) were shown to be predictive for errors \cite{ridderinkhof2003errors}. The mPFC has further been linked to cognitive control dynamics during action monitoring \cite{cavanagh2009prelude} in a flanker task. The motor cortex is further involved in predictive coding of future movements \cite{shipp2013reflections}, and it has been shown that premotor and motor cortices encode expected rewards \cite{ramkumar2016premotor}. Moreover, activity in the mPFC, the prefrontal cortex and motor cortex might serve as a top-down control signal that inhibits inappropriate responding \cite{narayanan2006top}.

Thus, our results fit very well in the involvement of frontal and precentral brain regions in action control and preparatory motor planning.

\section{Conclusion}

Here we have shown that deep learning methods, especially deep convolutional neural networks including residual neural networks, are not only among the best available machine learning methods for various decoding problems, but are also an invaluable tool for brain mapping, as shown in human intracranial recordings during error processing. By using sliding-window short-time decoding on single channels, we characterized the intracranial error response in depth; in comparison with the visualization of learned attributes in all-channel decoding, we could further show that both methods reveal overlapping, but not similar information, hinting towards the ability of CNNs to extract hidden connectivity features.

Deep convolutional networks were very accurate in all-channel decoding of errors from intracranial EEG electrodes in epilepsy patients, even though many of the channels were not informative. More compact or shallower networks, however, were prone to overfitting and not able to solve this challenging task reliably. For the intracranial classification of errors, we thus would recommend the use of at least 4-layered CNNs to reach an adequate number of parameters and thereby sufficient ability to learn.

\section*{Acknowledgment}

The authors would like to thank Pavel Kršek, Martin Tomášek, Peter C. Reinacher and Volker A. Coenen for their support, and the involved patients for their participation.

\bibliographystyle{IEEEtran}
\bibliography{IEEEabrv,main}

\end{document}